\documentclass{article}
\usepackage{spconf,amsmath,graphicx,multicol}
\usepackage[utf8]{inputenc}
\usepackage{amssymb}
\usepackage{amsmath}
\usepackage{graphics}
\usepackage{epsfig}
\usepackage{hyperref}
\usepackage{algorithm}
\usepackage[noend]{algpseudocode}

\usepackage{tikz}

\title{Technical Report of the Video Event Reconstruction and Analysis (VERA) System - Shooter Localization, Models, Interface, and Beyond}
\name{Junwei Liang$^{\star}$ \qquad  Jay D. Aronson$^{\dagger}$ \qquad  Alexander Hauptmann$^{\star}$}
\address{$^{\star}$\{junweil, alex\}@cs.cmu.edu \qquad
        $^{\dagger}$aronson@andrew.cmu.edu \\
        $^{\star}$School of Computer Science, Carnegie Mellon University \\
        $^{\dagger}$Center for Human Rights Science, Carnegie Mellon University }

\begin{document}
%
\maketitle
\begin{abstract}
Every minute, hundreds of hours of video are uploaded to social media sites and the Internet from around the world. This material creates a visual record of the experiences of a significant percentage of humanity and can help illuminate how we live in the present moment. When properly analyzed, this video can also help analysts to reconstruct events of interest, including war crimes, human rights violations, and terrorist acts. Machine learning and computer vision can play a crucial role in this process. 
In this technical report, we describe the Video Event Reconstruction and Analysis (VERA) system. This new tool brings together a variety of capabilities we have developed over the past few years (including video synchronization and geolocation to order unstructured videos lacking metadata over time and space, and sound recognition algorithms) to enable the reconstruction and analysis of events captured on video. Among other uses, VERA enables the localization of a shooter from just a few videos that include the sound of gunshots.
To demonstrate the efficacy of this suite of tools, we present the results of estimating the shooter's location of the Las Vegas Shooting in 2017 and show that VERA accurately predicts the shooter's location using only the first few gunshots. 
We then point out future directions that can help improve the system and further reduce unnecessary human labor in the process.
All of the components of VERA run through a web interface that enables human-in-the-loop verification to ensure accurate estimations. 
All relevant source code, including the web interface and machine learning models, is freely available on Github\footnote{https://vera.cs.cmu.edu}. 
We hope that researchers and software developers will be inspired to improve and expand this system moving forward to better meet the needs of human rights and public safety.

\end{abstract}

\section{Introduction}
\label{sec:intro}
With the ubiquity of high quality camera phones around the world, public events can now be captured and shared via social media instantly.
In a big public event with a large crowd of people, video recordings capture different moments of the event at different positions from different perspectives. 
These videos provide important information for analysts and researchers when crimes, including war crimes, terrorist acts or human rights violations, occur. 
The Center for Human Rights Science, for instance, has done video analysis as part of investigations in Syria, Ukraine, Nigeria, and Israel/Palestine~\cite{aronson2018utility, aronson2018reconstructing}, and is currently carrying out work on shootings in the United States. 
\begin{figure}
	\centering
		\includegraphics[width=0.47\textwidth]{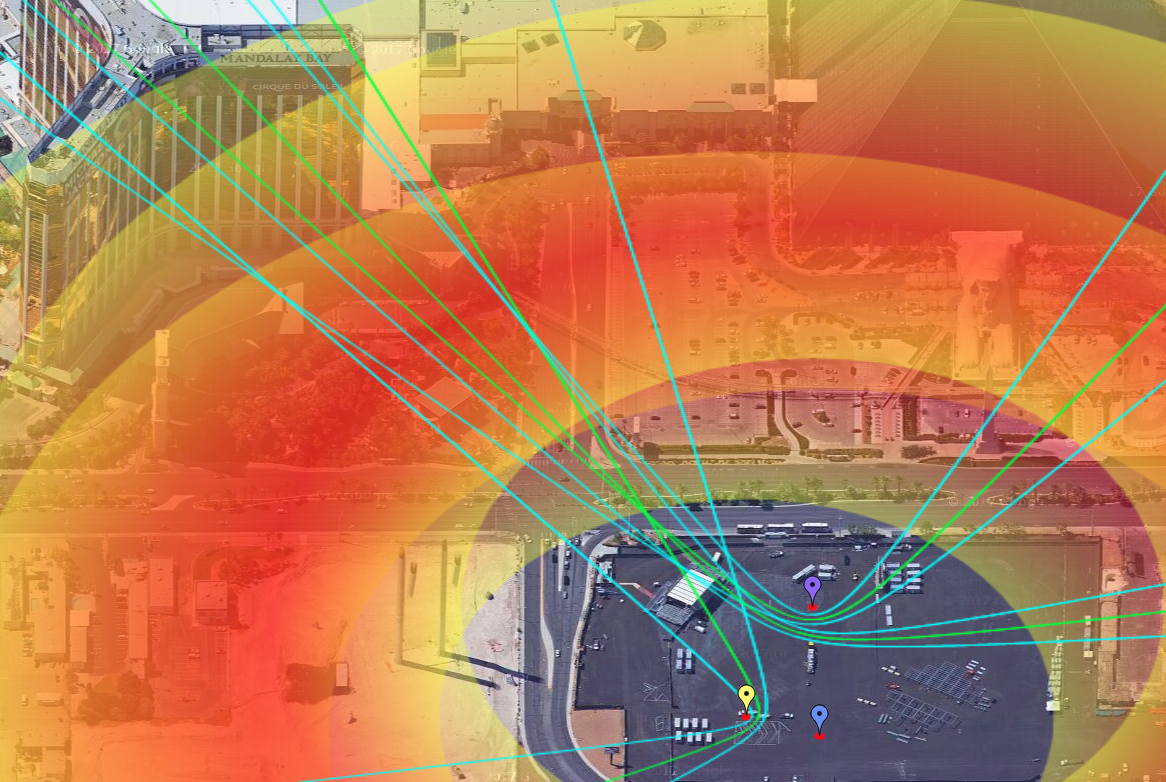}
	\caption{Visualization results of shooter localization using the VERA system for Las Vegas Shooting in 2017. It is computed based on only \textbf{three video recordings} as marked on the map and \textbf{single gunshot}. The red and yellow donut is the heatmap probability of the shooter location in \textbf{horizontal distance}. VERA also estimates that the shooter is likely to be within the light blue hyperbola lines. As we see, the overlapping area of all estimations points to the shooter's actual location - the north wing of the Mandalay Bay Hotel.}
	\label{daisy1}
	\vspace{-4mm}
\end{figure}
We have collected publicly available videos from widely publicized events in the United States, including the 2013 Boston Marathon bombing~\cite{chen2016videos} and the 2017 Las Vegas outdoor concert shooting. With these datasets, we can apply the tools we have developed for video analysis, including: semantic concept detection~\cite{liang2014semantic,liang2015detecting,liang2016learning,liang2017temporal,liang2017webly,liang2017leveraging,hinami2018multimodal,liang2016informedia,liang2016exploiting}, video captioning~\cite{jin2016generating,jin2016video}, intelligent question answering~\cite{liang2018focal,jiang2017memexqa,liang2019focal}, 3D event reconstruction~\cite{chen2016videos,liang2017event}, and activity prediction~\cite{liang2019peeking}. 
In addition to supporting the work of human rights activists, advocates, and investigators, we hope that these tools can help society achieve a better understanding of complex and chaotic events that cause social division and even violence~\cite{aronson2015video,liang2016video}.

Eyewitness videos are captured ``in the wild," and technical metadata is usually stripped once they are uploaded to social media.
These videos are often low quality and difficult to interpret when examined in isolation. 
Analysts generally need to go over a large number of these videos as useful information about the event may spread across different segments of different videos. 
We built the Video Event Reconstruction and Analysis (VERA) system, which is enabled by machine learning techniques like video synchronization~\cite{liang2017synchronization} and automated sound detection~\cite{liang2017temporal}, to solve this problem.

\begin{figure}
	\centering
		\includegraphics[width=0.47\textwidth]{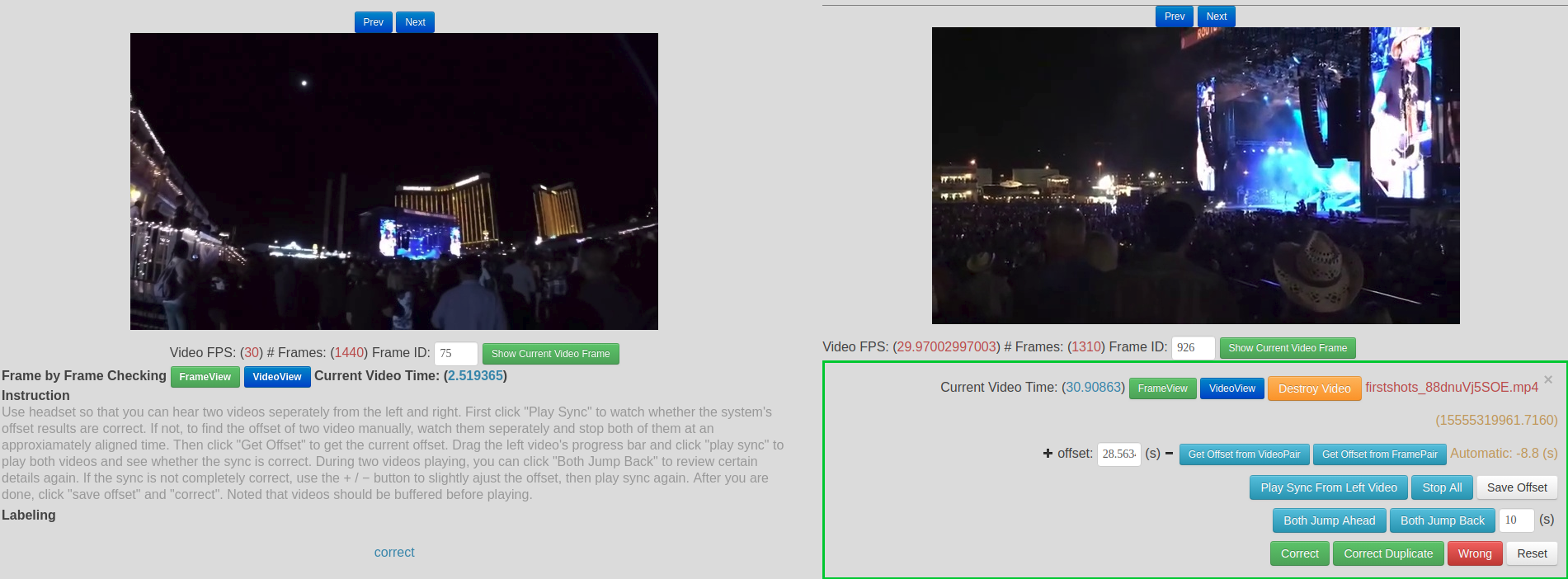}
	\caption{Pairwise video synchronization interface.}
	\label{fig:sync_interface}
	\vspace{-4mm}
\end{figure}

In this paper, we present how the VERA system can geolocate the shooter given a few video recordings that only capture the gunshot sound, but not necessarily an image of the shooter, as shown in Figure~\ref{daisy1}. 
The first step of the process utilizes semi-automated video synchronization~\cite{liang2017synchronization}, including a web interface for manual refinements, to put all unstructured videos into a global timeline. 
The VERA user then performs automatic gunshot detection~\cite{liang2017temporal} to temporally localize the gunshot segments within each video. Based on supersonic bullet physics~\cite{maher2008deciphering} and the physics of sound (described below), VERA identifies the exact moment in each video when the shockwave sound and muzzle blast sound (the two main aural components of a gunshot) occur, then computes the possible distances and directions of the shooter from each of the cameras. After putting each video on the map, VERA visualizes the potential range and direction of the shooter's location. 

All relevant source code, including the web interface and machine learning models, is freely available in Github. 
We do so in the hope that this tool can be used by anyone who needs it to protect and promote human rights and public safety. 
We also hope that researchers and software developers will be inspired to improve and expand this system moving forward to better meet the needs of the human rights community.
This report explicitly points out all major assumptions and likely sources of error so that users are informed and can take appropriate precautions when using the system or working to improve it.

This paper is organized as follows. In Section 2, we briefly describe our previously developed video synchronization methodology. In Section 3, we describe the gunshot detection system and how we can estimate shooter location based on gunshot sound. In Section 4, we explain our system architecture in terms of how the web interface interacts with back-end machine learning models. In Section 5, we discuss how VERA can be improved moving forward.

\section{Video Synchronization}
\label{sec:sync}
\begin{figure}
	\centering
		\includegraphics[width=0.47\textwidth]{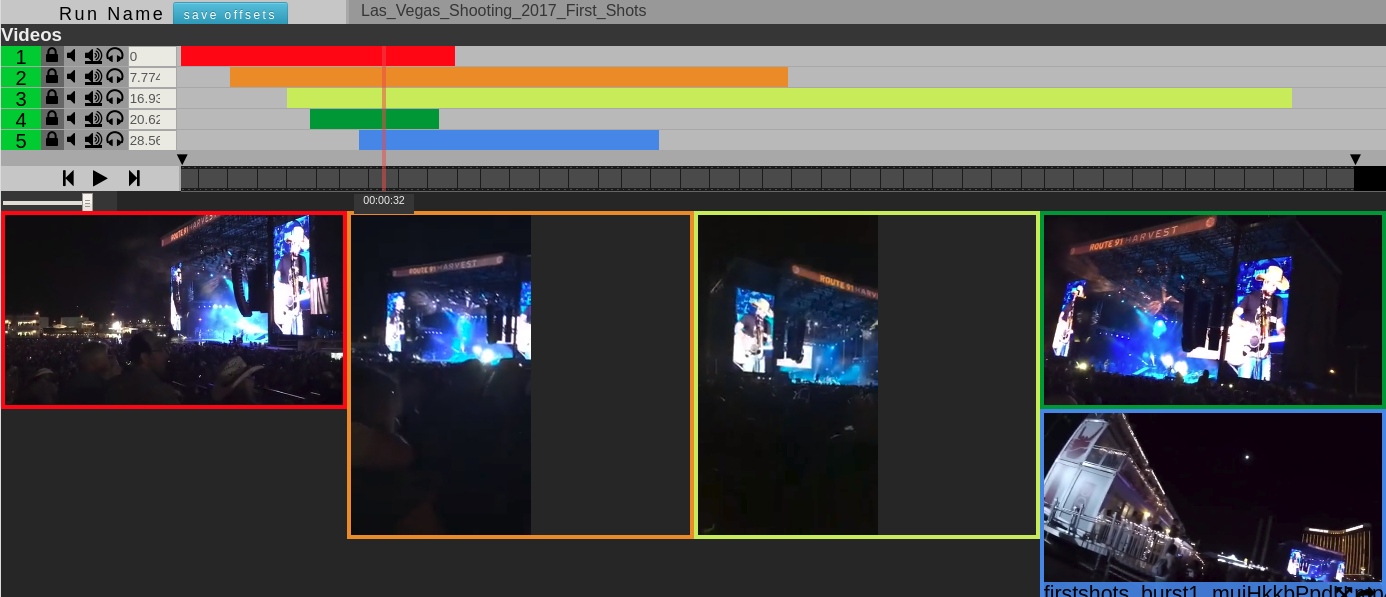}
	\caption{Video synchronization into a global timeline.}
	\label{fig:sync_global}
	\vspace{-4mm}
\end{figure}

Videos collected from social media are almost always stripped of metadata like global time stamps or GPS data that we could otherwise use to establish the time and place it was taken. 
Therefore, we need to start by putting the videos into a global timeline. 
After the user uploads all relevant videos in a collection to VERA, an automatic video synchronization model is utilized to organize the videos. Currently, the automatic system synchronizes the videos using sound. Please refer to ~\cite{liang2017synchronization} for more technical details. 
However, in order to use these videos for shooter localization, the synchronization needs to be refined using visual cues, as light travels much faster than sound from source to camera. VERA provides an easy-to-use interface as shown in Figure~\ref{fig:sync_interface}, which enables users to manually verify and synchronize pairs of videos down to the frame-level. 
Assuming users match the video pair to the exact frame pairs and the video FPS is 30, the error margin of the synchronization is within 33 milliseconds. The pairwise synchronization results are aggregated automatically into global results as shown in Figure~\ref{fig:sync_global}. Users can play the videos in a global timeline to understand the events in a coherent manner.

\section{Gunshot Localization}
\label{sec:sound}
\begin{figure}
	\centering
		\includegraphics[width=0.2\textwidth]{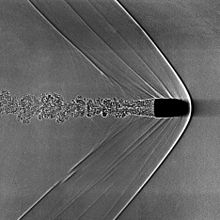}
	\caption{A shadowgraph of a supersonic bullet. Taken from wikipedia.}
	\label{fig:shadowgraph}
	\vspace{-4mm}
\end{figure}
\subsection{Overview}
Once all the videos with gunshots are placed into a global timeline, we can begin to estimate the distances of the shooter from each of the cameras if the bullet is supersonic (i.e., exceeding the speed of sound).
When a gunshots are captured in two videos, we can also estimate the direction of the bullet and location of the shooter based on the time differences of the muzzle blast sound~\footnote{The sound when the bullet exits the barrel of a gun} reaching each camera.
These estimations are computed for each video (Method 1, which is described in the next section) and each pair of videos (Method 2, which is described in Section 3.3), for a single gunshot. 
Each estimation provides an area of possible locations and the area where all estimations overlap is the most likely location of the shooter.
Users can repeat the calculations for multiple gunshots to get even more accurate localization results, or to trace the potential movements of the shooter over time and space.

\begin{figure}
	\centering
		\includegraphics[width=0.4\textwidth]{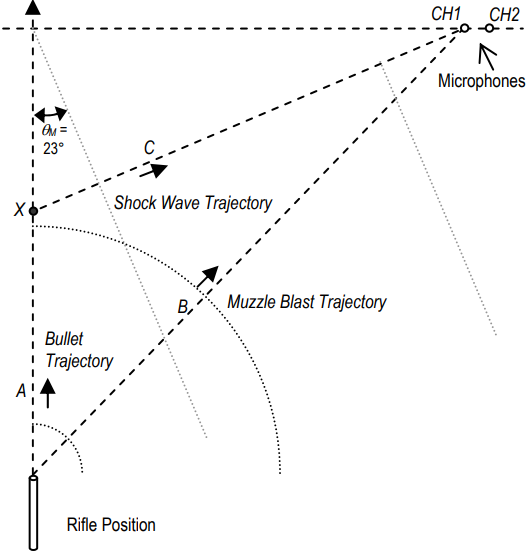}
	\caption{The physics model of how the shockwave sound and muzzle blast sound of a supersonic bullet reach the camera. Taken from ~\cite{maher2008deciphering}.}
	\label{fig:supersonic}
	\vspace{-4mm}
\end{figure}

\subsection{Method 1: Calculating Distance from the Camera in a Single Video }
\label{sec:method1}
This method requires the bullet to be supersonic. The main idea is that a supersonic bullet creates two distinctive sounds: a shockwave~\footnote{The sound of air getting compressed by a supersonic bullet. See more explanations at~\url{https://en.wikipedia.org/wiki/Bullet_bow_shockwave}} sound a muzzle blast sound. See here~\footnote{\url{https://vera.cs.cmu.edu/documents/method1.gif}} for an animation of the two sounds reaching a camera. If identified temporally, one can use the time difference of the two sound reaching the camera to estimate the distance between the camera and where the bullet is fired from. If the bullet is not supersonic, or the two sounds are not clearly distinguishable in the video (e.g., because of background noise or poor sound quality), the method described here will not work.

\begin{figure}
	\centering
		\includegraphics[width=0.47\textwidth]{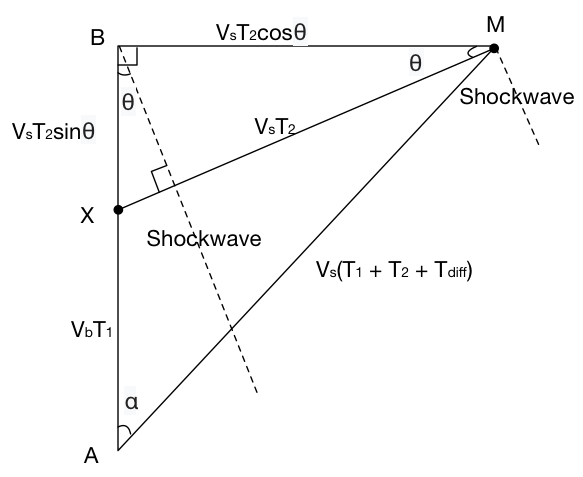}
	\caption{Method 1 math notation.}
	\label{fig:method1}
	\vspace{-4mm}
\end{figure}

Figure~\ref{fig:shadowgraph} shows a shadowgraph of a supersonic bullet. It depicts what happens to air when a bullet is travelling though it beyond the speed of sound. As we see, the bullet creates a cone-like shockwave wall that expands as the bullet travels. When this wall arrives the camera, it records the shockwave sound. The physics model of how the shockwave sound and muzzle blast sound of a supersonic bullet reach the camera is shown in Figure~\ref{fig:supersonic}. 
From the physics model, we can know that the bullet has to be coming in the camera's direction for the shockwave sound to be heard. 
For more details, please refer to ~\cite{maher2008deciphering}.

\begin{figure}
	\centering
		\includegraphics[width=0.47\textwidth]{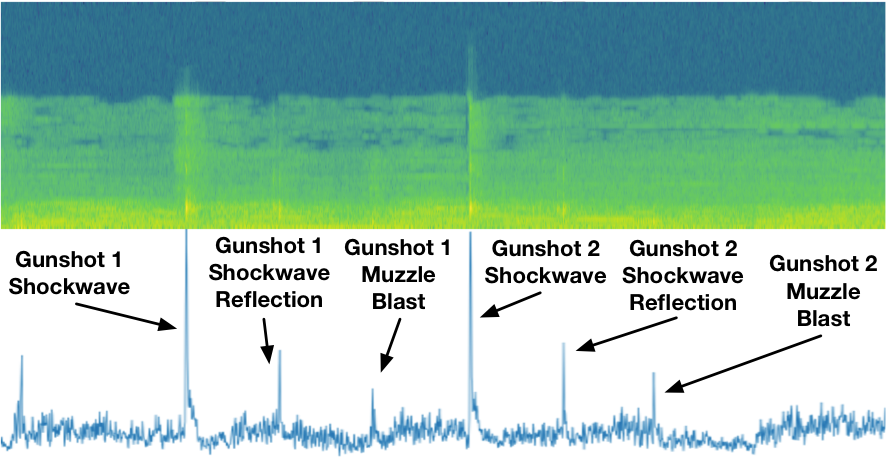}
	\caption{Example of gunshot spectrogram and power graph on the VERA interface. The example video ID is a-fCJGgqoxs on Youtube and the two gunshots showing in the graph are the 5th and 6th gunshots from the beginning. We can clearly see the power peaks and the users can mark them accordingly. However, sometimes the power graph for gunshots is not as clear as this one, especially when there are a lot of other loud noises.}
	\label{fig:spectrogram}
	\vspace{-4mm}
\end{figure}

\begin{figure}
	\centering
		\includegraphics[width=0.47\textwidth]{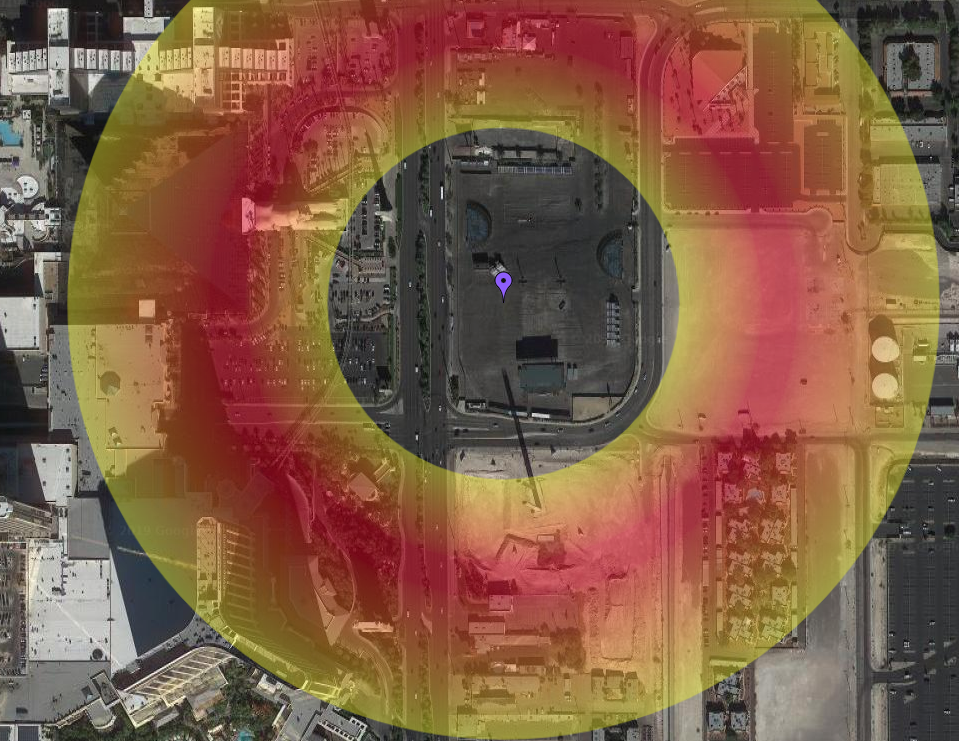}
	\caption{Example shooter localization using Method 1.}
	\label{donut}
	\vspace{-4mm}
\end{figure}

Based on the physics model, we can derive the computation graph as in Figure~\ref{fig:method1}. 
Suppose $V_s$ is the speed of sound, $V_b$ is the speed of the bullet, $\alpha$ is the angle between the camera to the shooter and the bullet trajectory. 
In reality, it is hard to know the exact speed of a bullet but we can learn the speed/velocity range of the bullet via existing field tests~\cite{maher2008deciphering} of that type of bullet~\footnote{\url{https://en.wikipedia.org/wiki/7.62\%C3\%9751mm_NATO}}.
$T_{diff}$ is the time difference between the camera records the shockwave sound and the muzzle blast sound. For the camera to record the shockwave sound, after the bullet is fired, the bullet travels $T_1$ under $V_b$ to point X, and then the shockwave travels at speed of sound $V_s$ for time $T_2$ to reach the camera. So:
\begin{equation}
\begin{split}
AB &= V_bT_1 + V_sT_2sin\theta = V_s(T_1 + T_2 + T_{diff})cos\alpha \\
BM &= V_sT_2sin\theta = V_s(T_1 + T_2 + T_{diff})sin\alpha
\end{split}
\end{equation}
Hence, we have:
\begin{equation}
\label{eq:method1}
\begin{split}
(V_b - V_scos\alpha)T_1 + (V_ssin\theta - V_scos\alpha)T_2 &= V_sT_{diff}cos\alpha \\
(-V_ssin\theta)T_1 + (V_ssin\theta - V_ssin\alpha)T_2 &= V_sT_{diff}sin\alpha
\end{split}
\end{equation}
In Eq~\ref{eq:method1}, the unknown variables are $T_1$ and $T_2$. Based on the two equations we can solve $T_1$ and $T_2$. Then the distance from the camera to the shooter, i.e. AM, can be computed by $V_s(T_1 + T_2 + T_{diff})$.
Currently in VERA, we ask users to input the range of the speed of sound $V_s$, the speed of the bullet $V_b$ and the angle $\alpha$. In practice the range of $\alpha$ is usually set from zero to fifteen, which already covers 30 degrees of freedom since the graph could be flipped. $\theta$ is given by $arcsin(V_s / V_b)$ according to ~\cite{maher2008deciphering}. $T_{diff}$ is currently marked by the users, aided by a spectrogram of the gunshot sound in VERA as shown in Figure~\ref{fig:spectrogram}.
Since we have a range of $V_s$, $V_b$ and $\alpha$, we use the Monte Carlo method (random sampling) to uniformly sample a value for each variables at a time to get $T_1$ and $T_2$, repeat many times (10k for example) and then report back the minimum, maximum and mean of the distance $D$. Please refer to the code for more details.
Note that the distance is the direct distance from the camera to the shooter. 
In order to have accurate visualization on the map which requires \textbf{horizontal} distance $D_h$, users can enter the elevation of the shooter $D_e$, and the horizontal distance is computed by $D_h = \sqrt{D^2 - {D_e}^2}$. 
Hence we get a donut-like possible area of the shooter for each video as shown in Figure~\ref{donut}. In future work, we will add 3D visualization in which the donut becomes a hollow ball so that possible elevated locations can also be shown.
In future work, we could automate the process of getting $V_b$ by automatic gun type detection based on the gunshot sound. It is a challenging problem as shown in previous work~\cite{maher2007acoustical,maher2008deciphering,maher2016wideband} since many types of guns make similar sounds. The type of bullet will also affect the characteristics of the gunshot sound even from the same gun, and environmental sound may complicate the audio recordings.
We could estimate the speed of sound if we know the air temperature of the event location~\cite{dean1982atmospheric}.

It is important to note that we assume the bullet travels at constant speed $V_b$ until the camera captures the shockwave sound. 
In reality, the bullet will lose speed as it travels due to drag (air resistance) and other factors. Indeed, the bullet speed may drop by half after traveling for 700 meters under certain conditions~\cite{maher2008deciphering}. 
We recommend users include a wide range of bullet speeds to compensate for this problem~\footnote{Currently, we are not sure about how much the bullet speed drop would affect our estimations and nor do we know how wide the range of bullet speeds would eliminate this effect.}, keeping in mind that the localization of the shooter becomes less certain as the range gets wider.

\subsection{Method 2: Geolocating a Shooter Using Two Videos}
\label{sec:method2}
\begin{figure}
	\centering
		\includegraphics[width=0.3\textwidth]{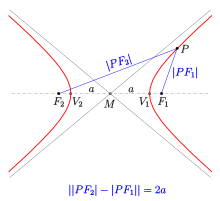}
	\caption{Hyperbola math notation. Taken from wikipedia.}
	\label{fig:hyperbola}
	\vspace{-4mm}
\end{figure}

This method applies when a pair of videos capture the muzzle blast of a single gunshot (note that this method can be applied to any distinct sound, such as an explosion). 
Method 2 makes use of the definition of a hyperbola as shown in Figure~\ref{fig:hyperbola}. The points (P) anywhere on the hyperbola satisfy that $||PF_2| - |PF_1|| = 2a$, and satisfy $|PF_2| - |PF_1| = 2a$ if we only consider points on the right part of the hyperbola.
In the gunshot localization case, for each pair of videos, since we have synchronized them and mark the muzzle blast sound in the videos' timeline, we know the time difference between camera 1 and camera 2 hearing the gunshot $T_{diff}$. Given the speed of sound $V_s$, we can compute the value of $2a = V_sT_{diff}$. After we put the two video camera locations on the map ($F_1$ and $F_2$), we can draw a hyperbola, which is likely to contain the shooter's location.

\begin{figure}
	\centering
		\includegraphics[width=0.47\textwidth]{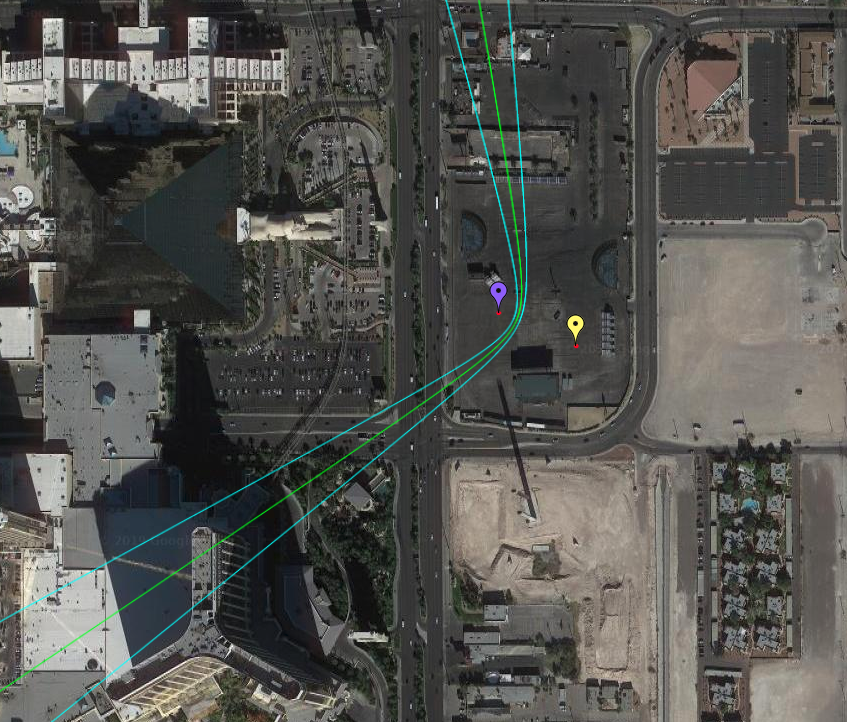}
	\caption{Example shooter localization using Method 2.}
	\label{fig:method2}
	\vspace{-4mm}
\end{figure}

As shown in Figure~\ref{fig:method2}, we can see three hyperbola lines. Recall in Section~\ref{sec:sync} we mention that the error margin of the video synchronization is 33 milliseconds, given that the frame matching could be off by half a frame. Also, users must enter the range of the speed of sound. The speed of sound is decided by the temperature of air and the changes of speed of sound vary in a small range as the temperature changes (the speed of sound is 331.3 m/s when it is 0 degree Celsius and it is 346 m/s when it is 25 degree Celsius~\footnote{~\url{https://en.wikipedia.org/wiki/Speed_of_sound}}).
Therefore we draw three hyperbola lines using the minimum, mean and maximum of the speed of sound. So we have three different value of $2a$: $V_{s_{min}}(T_{diff}-0.033)$, $(V_{s_{min}} + V_{s_{max}})(T_{diff}-0.033)/2.0$ and $V_{s_{max}}(T_{diff}+0.033)$. The second hyperbola is green colored while the others are light blue. The shooter has a high probability of being within the light blue lines, with the most highest probability being locations are on the green line.

To sum up, method 2 relies on accurate video synchronization, camera locations, and markings of the muzzle blast sound.

\subsection{Camera Locations}
Currently in VERA, we provide a Google Map interface and ask the users to \textbf{manually} mark the camera locations \textbf{at the time of} the video hearing the gunshots. The estimation of shooter's location would change drastically for method 2 (exponentially if the two cameras are close), and mildly for method 1, if the camera locations are different than marked. Users can see the different estimations by dragging the camera location markers on the web interface.
In future work, we expect to utilize Google Street View images or 45 degree view images to automatically match video frames to a GPS, enabling the determination of camera locations with less human labor.

\section{VERA Architecture}
\begin{figure}
	\centering
		\includegraphics[width=0.47\textwidth]{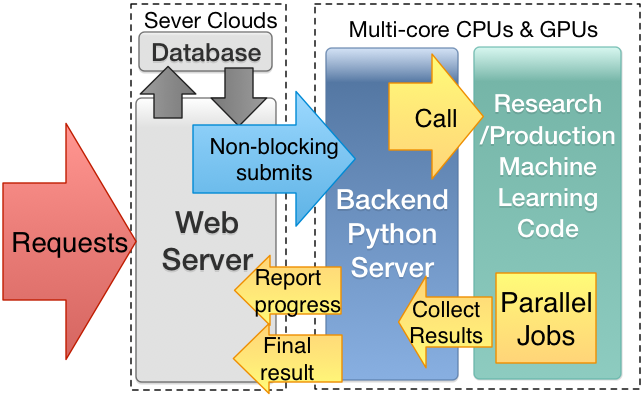}
	\caption{System architecture of VERA.}
	\label{fig:arch}
	\vspace{-4mm}
\end{figure}

We utilize production-ready web server - Apache server for serving the web requests and a flexible back-end Python server to leverage multi-CPU and multi-GPU computing cluster. Overall system architecture is show in Fig~\ref{fig:arch}. Future researchers could plug their machine learning components into the system seamlessly and efficiently.
\subsection{Web Interface}
The web interface system is password protected. Users need to register to get an account. VERA has a multi-level user system design, where the level 1 users, the administrators, will not be able to access the shooter localization tools but only a user management interface. They can delete or add level 2 users. Level 2 users are the ones created by admins or registered by the users themselves. Upon logging in, they will be directed to the shooter localization tool, where the main page shows that after creating the video collection (i.e., uploading all the videos), there are three steps for shooter localization: 1. Video synchronization; 2. Gunshot Blast/Shockwave Marking; 3. Shooter Localization per Gunshot. VERA uses AJAX at the browser side to communicate with the server. For more detailed and up-to-date description, please refer to the github site.

\subsection{Back-end Processing}
To ensure production efficiency, the web application code for our Apache server is written using PHP. However, our machine learning code usually runs on Python. Therefore we built a PHP-to-Python communication algorithm via sockets. Aside from a running Apache server, a multi-threading Python server is also running and monitors the PHP code. When a machine learning request initiates from the browser and reaches the PHP server via an AJAX POST request, the PHP server will send all required data to the Python server via socket. This request is \textbf{asynchronous}, meaning the PHP server will close the socket as soon as the data is sent.
When the Python server receives the request, it will spawn a dedicated thread to process the request. When the process is completed, the Python server will send back results to the PHP server via \textbf{Https} request. There is dedicated PHP code to process those results and put them into the database. Usually the back-end request involves running machine learning models, which may take a while, so the Python server will also send back progress updates using the same mechanism to tell the PHP server how much work is left. The browser side uses AJAX POST to ask the PHP server how much work is left until the task is complete.
The communication between the Python server and the PHP server is encrypted. Neither server will respond if the secret key doesn't match.
For more detailed and up-to-date description, please refer to the github site.

\section{Future Work and Conclusion}
As mentioned in each section, currently VERA has many parts that require human judgment and input. 
Moving forward, we plan to automate as much of the work requiring human action as possible, while still allowing human oversight and intervention as needed.
In this technical report, we showed how the VERA system can be used for shooter localization. We demonstrate how software engineering combined with machine learning and physics models can help localize shooters from unstructured social media videos. We demonstrate that with only three videos, we are able to correctly localize the shooter location in the case of the 2017 Las Vegas shooting. 
We are convinced that analysts and software developers will find ways of enhancing and reconfiguring the components of VERA for improved shooter localization and other tasks crucial to the protection and promotion of human rights and public safety. 

\noindent\textbf{Acknowledgements} 
This work was partially supported by the financial assistance award 60NANB17D156 from U.S. Department of Commerce, National Institute
of Standards and Technology (NIST). The U.S. Government is authorized to reproduce and distribute reprints for Governmental purposes notwithstanding any copyright annotation/herein. Disclaimer: The views and conclusions contained herein are those of the authors and should not be interpreted as necessarily representing the official policies or endorsements, either expressed or implied, of NIST, DOI/IBC, or the U.S. Government. 



\bibliographystyle{IEEEbib}
\bibliography{refs}

\end{document}